\documentclass[runningheads]{llncs}
\usepackage[T1]{fontenc}
\usepackage{graphicx,verbatim,hyperref}
\usepackage{booktabs}
\usepackage{color, colortbl}
\usepackage{amsmath}
\usepackage{amsfonts}
\usepackage{multirow}
\usepackage{xcolor}
\definecolor{Gray}{gray}{0.9}
\begin{document}
\title{BrainMT: A Hybrid Mamba-Transformer Architecture for Modeling Long-Range Dependencies in Functional MRI Data}
\titlerunning{BrainMT}
\author{Arunkumar Kannan \inst{1} \and
Martin A. Lindquist \inst{2} \and
Brian Caffo \inst{2}}
\authorrunning{A. Kannan et al.}
%
\institute{Dept. of Electrical and Computer Engineering, Johns Hopkins University, Baltimore, USA \and Dept. of Biostatistics, Johns Hopkins University, Baltimore, USA\\
\email{akannan7@jhu.edu}}


\maketitle              
\begin{abstract}
Recent advances in deep learning have made it possible to predict phenotypic measures directly from functional magnetic resonance imaging (fMRI) brain volumes, sparking significant interest in the neuroimaging community. However, existing approaches, primarily based on convolutional neural networks or transformer architectures, often struggle to model the complex relationships inherent in fMRI data, limited by their inability to capture long-range spatial and temporal dependencies. To overcome these shortcomings, we introduce BrainMT, a novel hybrid framework designed to efficiently learn and integrate long-range spatiotemporal attributes in fMRI data. Our framework operates in two stages: (1) a bidirectional Mamba block with a temporal-first scanning mechanism to capture global temporal interactions in a computationally efficient manner; and (2) a transformer block leveraging self-attention to model global spatial relationships across the deep features processed by the Mamba block. Extensive experiments on two large-scale public datasets, UKBioBank and the Human Connectome Project, demonstrate that BrainMT achieves state-of-the-art performance on both classification (sex prediction) and regression (cognitive intelligence prediction) tasks, outperforming existing methods by a significant margin. Our code and implementation details will be made publicly available at this \href{https://github.com/arunkumar-kannan/BrainMT-fMRI}{\textcolor{magenta}{link}}.


\end{abstract}
\section{Introduction}

The human brain is a complex, dynamic system whose intrinsic spatiotemporal organization can be investigated through functional connectivity, often modeled using functional magnetic resonance imaging (fMRI) \cite{lindquist2024measuring} data. In recent years, there has been significant interest in building predictive models that explore the relationship between functional connectivity and cognition to enhance understanding of neurological and neuropsychiatric disorders \cite{kannan2025gaming,dsouza2021m,smith2023regression,khosla2019machine}.

Existing predictive approaches for fMRI data can be broadly categorized into correlation-based and voxel-based methods. Correlation-based methods typically employ a two-step process to address the high dimensionality of 4D spatiotemporal fMRI data (three spatial dimensions and time). First, researchers use anatomical or functional parcellations or projections to reduce data dimensionality, followed by the computation of a subject-specific functional connectivity measure (typically a matrix). Data-driven models operate on these lower-dimensional matrices to predict a target variable associated with a downstream task. In the past decade, deep learning tools have been used as feature selection techniques in this pipeline, leveraging architectures such as convolutional neural networks \cite{kawahara2017brainnetcnn,khosla2019ensemble}, transformers \cite{kan2022brain,nguyen2020attend}, and graph neural networks \cite{li2021braingnn}. However, these methods face two main drawbacks.  
1) \emph{Loss of spatial structure}: reducing 4D data to 2D matrices can discard important spatial information, especially when very-low dimensional parcellations are used \cite{li2020detecting}. 
2) \emph{Lack of consensus in parcellation strategies}: different parcellation schemes and connectivity metrics can yield varying model performance \cite{abraham2017deriving}, motivating the need for predictive models that learn directly from raw volumetric fMRI data.

To address these limitations, recent studies have proposed using a voxel-based framework. Here preprocessed, voxel-level fMRI data are used as input in end-to-end deep learning pipelines. For instance, \cite{malkiel2022self} proposed a transformer-centric architecture for predicting cognitive intelligence and classifying schizophrenia, while \cite{kim2023swift} developed a swin transformer-based framework for subject-specific phenotype prediction. Despite these advances, models are constrained by the quadratic complexity of transformers, forcing them to process only relatively short sequences of fMRI volumes (10 to 20 frames) and then aggregate predictions across multiple time-window batches. Given the relative sluggishness of the hemodynamics underlying fMRI signals \cite{lindquist2008statistical}, restricting the model to a few time frames may overlook predictive temporal dynamics. In fact, \cite{kim2023swift} shows that incorporating more time frames steadily improves model performance.

In this work, we propose BrainMT, a novel hybrid deep learning framework that addresses the above challenges by combining a bi-directional Mamba block with a global transformer module to capture the full spatiotemporal complexity of fMRI data. Drawing inspiration from recent Mamba architectures \cite{li2024videomamba,park2024videomamba,hatamizadeh2024mambavision}, we adopt a temporal-first scanning mechanism to efficiently handle extended fMRI sequences, mitigating the computational bottlenecks found in earlier approaches. This design enables BrainMT to preserve and model long-range temporal signals while leveraging a lightweight global transformer to learn spatial dependencies between brain regions. Through extensive experiments and ablation studies on two large-scale public datasets - UKBioBank (UKB) and the Human Connectome Project (HCP), we demonstrate that BrainMT outperforms existing methods and generalizes robustly across diverse tasks for improved phenotypic prediction in neuroimaging.

\section{Methods}

\begin{figure}
    \centering
    \includegraphics[width=\linewidth]{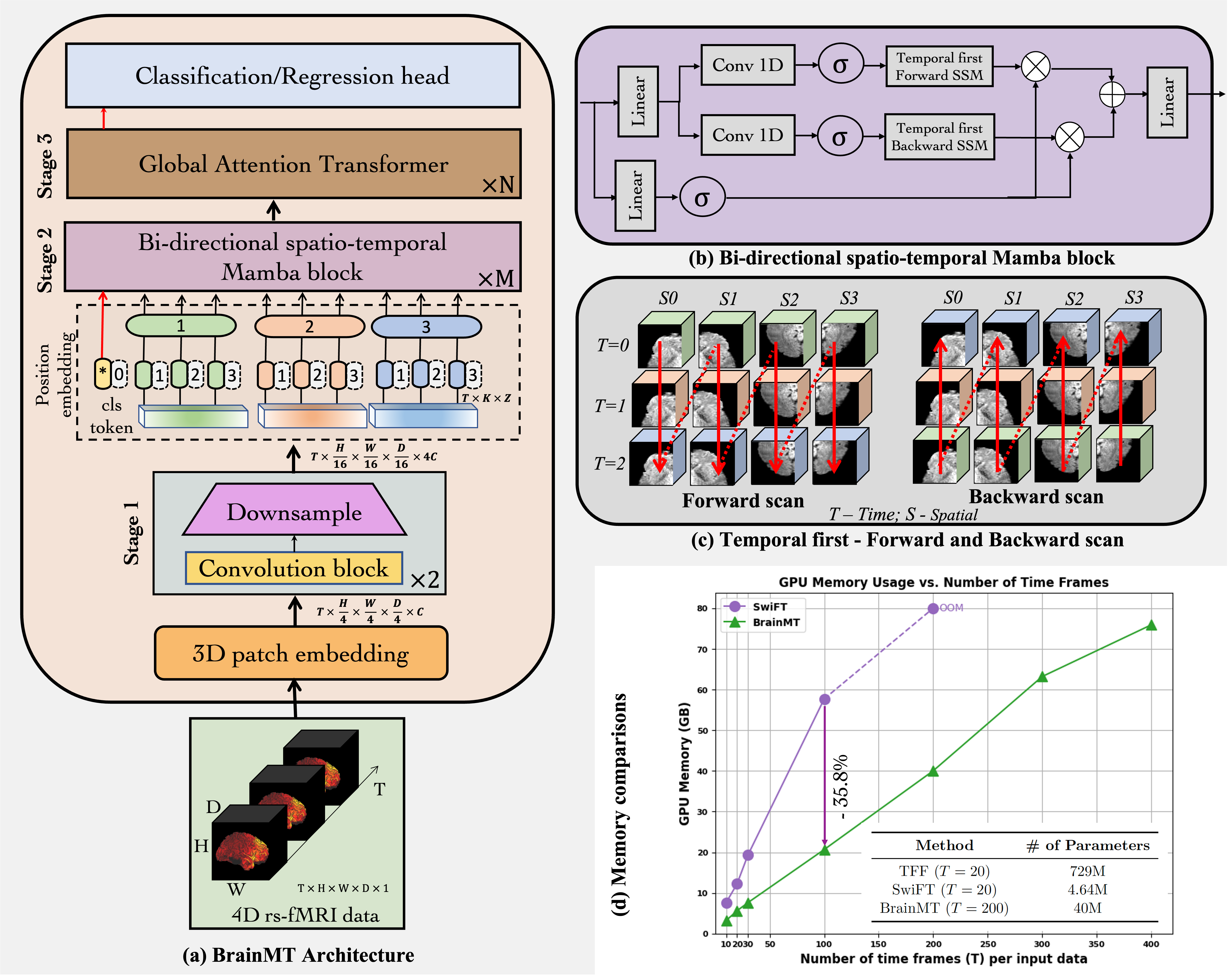}
    \caption{\textbf{Left}: (a) Overall architecture of BrainMT. \textbf{Right}: (b) Structure of the
bi-directional spatio-temporal Mamba block. (c) Scan mechanism. (d) Comparison of GPU memory usage and model parameters between BrainMT and SwiFT \cite{kim2023swift}.}
    \label{fig:teaser}
\end{figure}

Figure. \ref{fig:teaser} provides an overview of our proposed BrainMT framework, which employs a cascaded design comprising three primary components: (1) a convolution block; (2) a spatiotemporal Mamba block; and (3) a transformer block. Starting from fMRI data \(\mathbf{X} \in \mathbb{R}^{T \times H \times W \times D}\) - where \(T\) represents the number of sampled volumes (i.e., one per TR; the time resolution of the signal), and \(H\), \(W\), and \(D\) denote the three spatial dimensions - we first partition each volume into partially overlapping patches of size \(\frac{H}{4}\times\frac{W}{4}\times\frac{D}{4}\). These patches are then projected into a \(C\)-dimensional embedding space using two convolution layers. The resulting embedded feature maps serve as inputs to the following sub-modules.

\vspace{-1.5mm}

\noindent\textbf{Convolution block:} The convolution block acts as a hierarchical feature extractor for the embedded fMRI patches, generating multi-scale representations that capture coarse (high-resolution) and fine (low-resolution) details. In our framework, this block is implemented as a two-stage network, where each stage consists of a spatial convolution encoder followed by a downsampling operation. Formally, given embedded patches of size \(\frac{H}{4} \times \frac{W}{4} \times \frac{D}{4} \times C \), the convolution encoder outputs feature maps \textbf{F$_{i}$} with dimensions \(\frac{H}{4 \times 2^{i}} \times \frac{W}{4 \times 2^{i}} \times \frac{D}{4 \times 2^{i}}\)  and channel dimension \((C \times 2^{i}) \) for \(i \in \{1, 2\} \), using the following residual configuration:
\[
\begin{aligned}
\hat{\mathbf{X}} &= \texttt{GELU}(\texttt{LN}(\texttt{Conv3D}(\mathbf{X}_{\text{in}}))) \\
\mathbf{X}_{\text{conv}} &= \texttt{LN}(\texttt{Conv3D}(\hat{\mathbf{X}})) + \mathbf{X}_{\text{in}}
\end{aligned}
\]
where GELU is the Gaussian Error Linear Unit activation function and LN denotes layer normalization. \\

\noindent\textbf{Positional embedding:} Next, we add learnable positional embeddings \cite{arnab2021vivit,bertasius2021space} to the convolution block output, as the following Mamba block relies on token positions to model long-range dependencies. Let \(\mathbf{X}_{\text{conv}} \in \mathbb{R}^{T \times \frac{H}{16} \times \frac{W}{16} \times \frac{D}{16} \times Z}  \text{ with } Z = 4C,
\) be the convolution block’s output. First, each volumetric slice is flattened into a sequence of length \(
K = \frac{H}{16} \times \frac{W}{16} \times \frac{D}{16}.\) We then add two learnable positional embeddings: a spatial embedding 
\(\mathbf{P}_{s} \in \mathbb{R}^{1 \times K \times Z}\) and a temporal embedding 
\(\mathbf{P}_{t} \in \mathbb{R}^{T \times 1 \times Z}\) to the sequence. Additionally, we prepend a learnable classification token \(\mathbf{X}_{\text{cls}} \in \mathbb{R}^{1 \times 1 \times Z}\) to serve as a feature aggregator. This yields:
\[
\mathbf{X}_{\text{out}} = [\mathbf{X}_{\text{cls}},\, \mathbf{X}_{\text{conv}}] + \mathbf{P}_{s} + \mathbf{P}_{t},
\]
where \([\,\cdot,\,\cdot]\) denotes concatenation along the token dimension. Finally, we reshape \(\mathbf{X}_{\text{out}}\) from \(\mathbb{R}^{T \times K \times Z}\) to \(\mathbb{R}^{1 \times L \times Z}\) with \(L = T \times K\), providing a longer context sequence for subsequent spatiotemporal modeling. \\

\noindent\textbf{Spatio-temporal Mamba block:} The Mamba block employs bi-directional selective state-space models (SSMs) to capture long-context representations from the input sequence. In what follows, we first outline the fundamentals of SSMs and their discretization in Mamba, and then describe the architecture that adapts Mamba to 4D spatiotemporal data.

\noindent(i) \underline{\emph{SSM and Mamba preliminaries:}} State-space models (SSMs), which originate from Kalman filters, are linear-time-invariant (LTI) systems that map a one-dimensional continuous input sequence $x(t) \in \mathbb{R}$ to a one-dimensional output sequence $y(t) \in \mathbb{R}$ via an internal hidden state $h(t) \in \mathbb{R}^{N}$, where $N$ is the dimension of the hidden state. Formally,
\begin{align}
    h'(t) &= \mathbf{A} \, h(t) + \mathbf{B} \, x(t), \\
    y(t) &= \mathbf{C} \, h(t),
\end{align}
where $\mathbf{A} \in \mathbb{R}^{N \times N}$ is the evolution matrix, and $\mathbf{B} \in \mathbb{R}^{N \times 1}$ and $\mathbf{C} \in \mathbb{R}^{1 \times N}$ are projection matrices. In order to handle discrete input data (e.g., images or fMRI sequences), Mamba \cite{gu2023mamba} applies a Zero-Order Hold (ZOH) discretization to $\mathbf{A}$, $\mathbf{B}$, $\mathbf{C}$ using a time-scale parameter $\Delta$:
\begin{align}
    \overline{\mathbf{A}} &= \exp(\Delta \mathbf{A}), \quad
    \overline{\mathbf{B}} = (\Delta \mathbf{A})^{-1}\!\Bigl(\exp(\Delta \mathbf{A}) - \mathbf{I}\Bigr) \,\Delta \mathbf{B}, \\
    h(t) &= \overline{\mathbf{A}} \, h(t - 1) + \overline{\mathbf{B}} \, x(t), \quad
    y(t) = \mathbf{C} \, h(t).
\end{align}
Because SSMs are LTI, their parameters $\mathbf{A}$, $\mathbf{B}$, $\mathbf{C}$ remain fixed for all time steps, which can limit contextual learning. To overcome this, Mamba introduces a selective scan mechanism that make $\mathbf{B}$, $\mathbf{C}$, $\Delta$ input-dependent, thereby enabling the model to adaptively learn contextual relations from longer sequences.

\noindent(ii) \underline{\emph{Architecture:}} While the original Mamba formulation operates on 1D data (e.g., text), vision tasks require modeling spatial relationships. Hence, we adopt the Vision Mamba block \cite{zhu2024vision}, which incorporates bi-directional selective SSMs to capture spatiotemporal context in the data. Figure. \ref{fig:teaser}b provides a schematic illustration. The input sequence is first linearly projected into two vectors, $\mathbf{x}$ and $\mathbf{z}$, each of dimension $d$. To prevent tokens from being processed independently, a 1D convolution is applied to $\mathbf{x}$ before feeding it into forward and backward selective SSM modules, producing $\mathbf{y}_{\text{forward}}$ and $\mathbf{y}_{\text{backward}}$. Both outputs are then gated by $\mathbf{z}$ and summed to form the final output sequence. By processing the input in both directions and selectively adjusting the SSM parameters, the Vision Mamba block can learn rich spatiotemporal dependencies from fMRI data.

In addition, we adopt a \emph{temporal-first scanning mechanism} as shown in Figure. \ref{fig:teaser}c, where tokens are arranged so that time is treated as the leading dimension, followed by spatial dimensions. This design choice is motivated by neurobiological principles of coactivation in fMRI data, where consistent temporal correlations among brain regions are critical for identifying functional connectivity. By ensuring that the SSMs process the time dimension prior to spatial dimensions, the Vision Mamba block can more effectively leverage long-range temporal interactions alongside spatial dependencies from fMRI data. \\

\noindent\textbf{Transformer block:}
The final stage of our framework leverages a multi-head self-attention module to capture global dependencies. Let $\mathbf{Q}, \mathbf{K}, \mathbf{V} \in \mathbb{R}^{L \times Z}$ represent the query, key, and value matrices, where $L$ is the sequence length and $Z$ is the token dimensionality. The self-attention mechanism is:
\begin{equation}
    \mathrm{Attention}(\mathbf{Q}, \mathbf{K}, \mathbf{V}) 
    = \mathrm{Softmax}\!\Bigl(\frac{\mathbf{Q}\,\mathbf{K}^{T}}{\sqrt{d_{\mathrm{head}}}}\Bigr)\,\mathbf{V},
\end{equation}
where $d_{\mathrm{head}}$ is the dimension of each attention head. Although self-attention has a quadratic computational complexity in $L$, our initial convolution and downsampling stages significantly reduce the sequence length, thereby making the transformer block tractable for longer sequences.
\vspace{-1.5mm}
\section{Experimental Setup and Results}

\noindent\textbf{Datasets and preprocessing:} We applied our method to resting-state fMRI data from 6000 UKB participants \cite{sudlow2015uk} and 1075 HCP (S1200 release) participants \cite{smith2013resting}. We used the preprocessed data provided with both datasets \cite{glasser2013minimal,alfaro2018image}, which follows the ``fMRI volume'' pipeline (bias field reduction, skull stripping, cross-modality registration, and spatial normalization). To ensure stable network training, we applied global Z-score normalization, excluding background regions, which were filled with the minimum Z-score intensity \cite{malkiel2022self}. We split the data into 70\% for training, 15\% for validation, and 15\% for testing. Our prediction targets were sex and cognitive intelligence scores: specifically the ``cognitive function” composite scores from HCP and the ``fluid intelligence/reasoning” scores from UKB. All regression targets were Z-score normalized. In correlation-based approaches, we parcellate the data using the HCP multimodal atlas \cite{glasser2016multi} and compute Pearson correlations to capture functional connectivity between brain regions. \\

\noindent\textbf{Implementation details:} We implemented our model in PyTorch and trained it on NVIDIA L40S GPUs (48GB RAM). For all downstream tasks, we used the same BrainMT architecture with 2 convolution blocks, 12 Mamba blocks, and 8 transformer blocks in sequence. The final output is obtained by applying normalization to the \(\mathbf{X}_{\text{cls}}\) token, followed by an MLP head. We randomly selected 200 frames from resting-state fMRI as input (justified in Table \ref{tab:ablation}) and used default Mamba hyperparameters for the spatiotemporal layer (state dimension: 16, expansion ratio: 2). Training used a distributed data-parallel strategy with AdamW under a cosine learning rate schedule over 20 epochs, with the first 5 as a linear warm-up. Default values were used for learning rate (2e-4), weight decay (0.05), and batch size (2), with early stopping based on validation loss. All hyperparameters were fixed using the validation set. For intelligence prediction, we minimized mean squared error (MSE) and evaluated using MSE, mean absolute error (MAE), and Pearson’s R. For sex classification, we optimized binary cross-entropy loss and evaluated using balanced accuracy (B.Acc), standard accuracy (Acc.), and the area under the receiver operating characteristic curve (AUROC).

\begin{table}[ht]
\centering
\caption{Comparisons with baseline models on HCP and UKB Datasets for cognitive intelligence prediction. Values are given with their standard deviations. Lower MSE and MAE, and higher R indicate better performance. Color convention: \textcolor{red}{best}, \textcolor{blue}{2nd-best}.}
\label{tab:intelligence}
\resizebox{\textwidth}{!}{%
\ttfamily
\begin{tabular}{rccc|ccc}
 & \multicolumn{3}{c}{\textbf{HCP}} & \multicolumn{3}{c}{\textbf{UKBioBank}} \\
\cline{2-4} \cline{5-7}
\textbf{Method} 
 & \textbf{MSE}
 & \textbf{MAE}
 & \textbf{R}
 & \textbf{MSE}
 & \textbf{MAE}
 & \textbf{R}\\
\hline
XG-Boost ~\cite{chen2016xgboost}
 & 1.004 \scriptsize \textcolor{gray}{0.22}
 & 0.831 \scriptsize \textcolor{gray}{0.15}
 & 0.14 \scriptsize \textcolor{gray}{0.05}
 & 1.049 \scriptsize \textcolor{gray}{0.19}
 & 0.811 \scriptsize \textcolor{gray}{0.14}
 & 0.01 \scriptsize \textcolor{gray}{0.03} \\
BrainNetCNN \cite{kawahara2017brainnetcnn}
 & 0.981 \scriptsize \textcolor{gray}{0.21}
 & 0.799 \scriptsize \textcolor{gray}{0.10}
 & 0.21 \scriptsize \textcolor{gray}{0.02} 
 & 1.003 \scriptsize \textcolor{gray}{0.23}
 & 0.801 \scriptsize \textcolor{gray}{0.16}
 & 0.01 \scriptsize \textcolor{gray}{0.01}  \\

BrainGNN \cite{li2021braingnn}
 & 0.946 \scriptsize \textcolor{gray}{0.02}
 & 0.791 \scriptsize \textcolor{gray}{0.10}
 & 0.28 \scriptsize \textcolor{gray}{0.04} 
 & 0.995 \scriptsize \textcolor{gray}{0.13}
 & 0.794 \scriptsize \textcolor{gray}{0.03}
 & 0.06 \scriptsize \textcolor{gray}{0.10}  \\
BrainNetTF \cite{kan2022brain}
 & 0.998 \scriptsize \textcolor{gray}{0.17}
 & 0.820 \scriptsize \textcolor{gray}{0.03}
 & 0.18 \scriptsize \textcolor{gray}{0.10} 
 & 0.999 \scriptsize \textcolor{gray}{0.01}
 & 0.798 \scriptsize \textcolor{gray}{0.15}
 & 0.04 \scriptsize \textcolor{gray}{0.05}  \\
\hline
TFF \cite{malkiel2022self}
 & 0.957 \scriptsize \textcolor{gray}{0.10}
 & 0.798 \scriptsize \textcolor{gray}{0.02}
 & 0.27 \scriptsize \textcolor{gray}{0.08} 
 & 0.998 \scriptsize \textcolor{gray}{0.02}
 & 0.795 \scriptsize \textcolor{gray}{0.11}
 & 0.04 \scriptsize \textcolor{gray}{0.03} \\

SwiFT \cite{kim2023swift}
 & \textcolor{blue}{0.914} \scriptsize \textcolor{gray}{0.11}
 & \textcolor{blue}{0.790} \scriptsize \textcolor{gray}{0.04}
 & \textcolor{blue}{0.32} \scriptsize \textcolor{gray}{0.02}  
 & \textcolor{blue}{0.994} \scriptsize \textcolor{gray}{0.16}
 & \textcolor{blue}{0.791} \scriptsize \textcolor{gray}{0.03}
 & \textcolor{blue}{0.07} \scriptsize \textcolor{gray}{0.01} \\

BrainMT ~~~~
 & \textbf{\textcolor{red}{0.835}} \scriptsize \textcolor{gray}{0.02}
 & \textbf{\textcolor{red}{0.741}} \scriptsize \textcolor{gray}{0.01}
 & \textbf{\textcolor{red}{0.41}} \scriptsize \textcolor{gray}{0.03}
 & \textbf{\textcolor{red}{0.932}} \scriptsize \textcolor{gray}{0.03}
 & \textbf{\textcolor{red}{0.773}} \scriptsize \textcolor{gray}{0.01}
 & \textbf{\textcolor{red}{0.24}} \scriptsize \textcolor{gray}{0.02}\\
\hline

\end{tabular}%
\normalfont
}
\end{table}

\noindent\textbf{Quantitative results:} To evaluate the effectiveness of BrainMT, we performed a comprehensive analysis against state-of-the-art methods representing both correlation-based and voxel-based approaches. For the correlation-based methods, we compared BrainMT with XGBoost \cite{chen2016xgboost} (our standard machine learning baseline), BrainNetCNN \cite{kawahara2017brainnetcnn}, BrainGNN \cite{li2021braingnn}, and BrainNetTF \cite{kan2022brain}. For the voxel-based methods, we used TFF \cite{malkiel2022self} and SwiFT \cite{kim2023swift} as baselines. In each case, we followed the original studies’ hyperparameter settings and implementations. Quantitative results obtained from a repeated three-fold cross-validation are reported in Table. \ref{tab:intelligence} for cognitive intelligence prediction and in Table. \ref{tab:sex} for sex classification. On the intelligence prediction task, BrainMT consistently outperformed all baseline methods by a significant margin. Notably, most baseline methods yielded an MSE close to 1.0 in UKB (with targets normalized to have a variance of 1), suggesting that they mostly predict the sample mean. In contrast, BrainMT achieved a 6.23\% reduction in MSE on the UKB dataset and a 8.75\% reduction on the HCP dataset. For sex classification, BrainMT significantly outperformed all baselines on the HCP dataset and matched the performance of SwiFT on the UKB dataset. We further show that BrainMT is memory-efficient in Figure \ref{fig:teaser}d, where GPU memory usage and model parameters are compared with SwiFT across varying frame counts, demonstrating that BrainMT is 35.8\% more memory-efficient and maintains linear complexity in \(T\).

\begin{table}[t!]
\centering
\caption{Comparisons with baseline models for sex classification. Higher Acc., B.Acc, and AUROC indicate better performance. Color convention: \textcolor{red}{best}, \textcolor{blue}{2nd-best}}
\label{tab:sex}
\resizebox{\textwidth}{!}{%
\ttfamily
\begin{tabular}{rccc|ccc}
 & \multicolumn{3}{c}{\textbf{HCP}} & \multicolumn{3}{c}{\textbf{UKBioBank}} \\
\cline{2-4} \cline{5-7}
\textbf{Method} 
 & \textbf{Acc.}
 & \textbf{B.Acc}
 & \textbf{AUROC}
 & \textbf{Acc.}
 & \textbf{B.Acc}
 & \textbf{AUROC}\\
\hline

XG-Boost ~\cite{chen2016xgboost}
 & 68.43 \scriptsize \textcolor{gray}{2.37}
 & 67.86 \scriptsize \textcolor{gray}{3.35}
 & 73.2 \scriptsize \textcolor{gray}{2.43}
 & 79.15 \scriptsize \textcolor{gray}{1.38}
 & 78.27 \scriptsize \textcolor{gray}{1.37}
 & 86.3 \scriptsize \textcolor{gray}{0.42}\\

BrainNetCNN \cite{kawahara2017brainnetcnn}
 & 76.94 \scriptsize \textcolor{gray}{3.29}
 & 75.41 \scriptsize \textcolor{gray}{2.31}
 & 82.3 \scriptsize \textcolor{gray}{2.35}
 & 85.78 \scriptsize \textcolor{gray}{0.41}
 & 84.82 \scriptsize \textcolor{gray}{0.38}
 & 92.4 \scriptsize \textcolor{gray}{0.32} \\


BrainGNN \cite{li2021braingnn}
 & 84.73 \scriptsize \textcolor{gray}{1.22}
 & 84.26 \scriptsize \textcolor{gray}{0.98}
 & 90.1 \scriptsize \textcolor{gray}{ 0.73}
 & 89.85 \scriptsize \textcolor{gray}{1.28}
 & 89.71 \scriptsize \textcolor{gray}{0.21}
 & 95.8 \scriptsize \textcolor{gray}{1.14} \\

BrainNetTF \cite{kan2022brain}
 & 82.87 \scriptsize \textcolor{gray}{2.19}
 & 81.55 \scriptsize \textcolor{gray}{3.16}
 & 89.3 \scriptsize \textcolor{gray}{2.09}
 & 87.92 \scriptsize \textcolor{gray}{1.23}
 & 87.38 \scriptsize \textcolor{gray}{1.15}
 & 95.1 \scriptsize \textcolor{gray}{0.71}\\
\hline
TFF \cite{malkiel2022self}
 & 92.94 \scriptsize \textcolor{gray}{1.17}
 & 92.40 \scriptsize \textcolor{gray}{1.19}
 & 97.1 \scriptsize \textcolor{gray}{2.08}
 & 96.11 \scriptsize \textcolor{gray}{0.13}
 & 95.46 \scriptsize \textcolor{gray}{0.28}
 & 99.3 \scriptsize \textcolor{gray}{0.12}\\

SwiFT \cite{kim2023swift}
 & \textcolor{blue}{93.06} \scriptsize \textcolor{gray}{ 1.08}
 & \textcolor{blue}{92.61} \scriptsize \textcolor{gray}{ 1.26}
 & \textcolor{blue}{97.5} \scriptsize \textcolor{gray}{1.95}
 & \textcolor{blue}{97.45} \scriptsize \textcolor{gray}{0.11}
 & \textcolor{blue}{97.69} \scriptsize \textcolor{gray}{0.05}
 & \textbf{\textcolor{red}{99.4}} \scriptsize \textcolor{gray}{0.13} \\

BrainMT ~~~~
 & \textbf{\textcolor{red}{96.28}} \scriptsize \textcolor{gray}{0.02}
 & \textbf{\textcolor{red}{96.15}} \scriptsize \textcolor{gray}{0.03}
 & \textbf{\textcolor{red}{98.7}} \scriptsize \textcolor{gray}{0.06}
 & \textbf{\textcolor{red}{97.91}} \scriptsize \textcolor{gray}{0.09}
 & \textbf{\textcolor{red}{97.77}} \scriptsize \textcolor{gray}{0.08}
 & \textcolor{blue}{99.2} \scriptsize \textcolor{gray}{0.04} \\
\hline

\end{tabular}%
\normalfont
}
\end{table}

\begin{table}[ht]
\centering
\caption{Ablation studies on different number of time frames ($\mathcal{A}$), architecture configurations ($\mathcal{B}$), number of Mamba (M) and Transformer (N) layers ($\mathcal{C}$), visual mamba blocks ($\mathcal{D}$) and predicting functional connectivity correlations ($\mathcal{E}$).}
\label{tab:ablation}
\resizebox{\textwidth}{!}{%
\ttfamily
\begin{tabular}{ccccc|ccc}
 & & \multicolumn{3}{c}{\textbf{HCP}} & \multicolumn{3}{c}{\textbf{UKBioBank}} \\
\cline{3-5} \cline{6-8}
\textbf{Exp} & \textbf{Configuration} & \textbf{MSE} & \textbf{MAE} & \textbf{R} & \textbf{MSE} & \textbf{MAE} & \textbf{R} \\
\hline

\multirow{2}{*}{$\mathcal{A}$} 
 & $T$ = 100 
 & 0.861 \scriptsize \textcolor{gray}{0.19}
 & 0.765 \scriptsize \textcolor{gray}{0.14}
 & 0.38 \scriptsize \textcolor{gray}{0.11}
 & 0.966 \scriptsize \textcolor{gray}{0.12}
 & 0.783 \scriptsize \textcolor{gray}{0.08}
 & 0.20 \scriptsize \textcolor{gray}{0.05} \\ 
 & $T$ = 300 
 & 0.870 \scriptsize \textcolor{gray}{0.08}
 & 0.767 \scriptsize \textcolor{gray}{0.11}
 & 0.37 \scriptsize \textcolor{gray}{0.06}
 & 0.971 \scriptsize \textcolor{gray}{0.15}
 & 0.785 \scriptsize \textcolor{gray}{0.16}
 & 0.19 \scriptsize \textcolor{gray}{0.13}\\
\hline

\multirow{3}{*}{$\mathcal{B}$} 
 & No Transformer 
 & 0.865 \scriptsize \textcolor{gray}{0.02}
 & 0.766 \scriptsize \textcolor{gray}{0.13}
 & 0.37 \scriptsize \textcolor{gray}{0.05}
 & 0.962 \scriptsize \textcolor{gray}{0.16}
 & 0.782 \scriptsize \textcolor{gray}{0.07}
 & 0.20 \scriptsize \textcolor{gray}{0.01}\\
 & No Conv 
 & 0.849 \scriptsize \textcolor{gray}{0.17} 
 & 0.754 \scriptsize \textcolor{gray}{0.09}
 & 0.39 \scriptsize \textcolor{gray}{0.11}
 & 0.949 \scriptsize \textcolor{gray}{0.13}
 & 0.778 \scriptsize \textcolor{gray}{0.18}
 & 0.21 \scriptsize \textcolor{gray}{0.03} \\
 & No Conv \& Transf. 
 & 0.857 \scriptsize \textcolor{gray}{0.06} 
 & 0.759 \scriptsize \textcolor{gray}{0.15}
 & 0.38 \scriptsize \textcolor{gray}{0.04}
 & 0.955 \scriptsize \textcolor{gray}{0.15}
 & 0.779 \scriptsize \textcolor{gray}{0.12}
 & 0.20 \scriptsize \textcolor{gray}{0.08} \\
\hline

\multirow{2}{*}{$\mathcal{C}$} 
 & Large \scriptsize (24M, 16N) 
 & 0.874 \scriptsize \textcolor{gray}{0.04}
 & 0.768 \scriptsize \textcolor{gray}{0.07}
 & 0.37 \scriptsize \textcolor{gray}{0.08}
 & 0.951 \scriptsize \textcolor{gray}{0.01}
 & 0.780 \scriptsize \textcolor{gray}{0.08}
 & 0.21 \scriptsize \textcolor{gray}{0.05} \\ 
 & Small \scriptsize (6M, 4N) 
 & 0.883 \scriptsize \textcolor{gray}{0.14}
 & 0.772 \scriptsize \textcolor{gray}{0.10}
 & 0.36 \scriptsize \textcolor{gray}{0.16}
 & 0.965 \scriptsize \textcolor{gray}{0.04}
 & 0.783 \scriptsize \textcolor{gray}{0.07}
 & 0.20 \scriptsize \textcolor{gray}{0.09} \\
\hline

\multirow{2}{*}{$\mathcal{D}$} 
 & VMamba \cite{liu2024vmamba}
 & 0.891 \scriptsize \textcolor{gray}{0.38} 
 & 0.790 \scriptsize \textcolor{gray}{0.20}
 & 0.24 \scriptsize \textcolor{gray}{0.13}
 & 0.968 \scriptsize \textcolor{gray}{0.29} 
 & 0.783 \scriptsize \textcolor{gray}{0.28}
 & 0.15 \scriptsize \textcolor{gray}{0.19} \\
 & MambaVision \cite{hatamizadeh2024mambavision}
 & 0.913 \scriptsize \textcolor{gray}{0.21}
 & 0.801 \scriptsize \textcolor{gray}{0.25}
 & 0.07 \scriptsize \textcolor{gray}{0.01}
 & 0.979 \scriptsize \textcolor{gray}{0.10}
 & 0.787 \scriptsize \textcolor{gray}{0.23}
 & 0.09 \scriptsize \textcolor{gray}{0.18} \\
\hline

\multirow{2}{*}{$\mathcal{E}$} 
 & SwiFT 
 & 0.259 \scriptsize \textcolor{gray}{0.48}
 & 0.047 \scriptsize \textcolor{gray}{0.21}
 & 0.54 \scriptsize \textcolor{gray}{0.35}
 & 0.126 \scriptsize \textcolor{gray}{0.29}
 & 0.031 \scriptsize \textcolor{gray}{0.27}
 & 0.65 \scriptsize \textcolor{gray}{0.31} \\
 & BrainMT 
 & 0.187 \scriptsize \textcolor{gray}{0.38}
 & 0.035 \scriptsize \textcolor{gray}{0.25}
 & 0.63 \scriptsize \textcolor{gray}{0.14}
 & 0.083 \scriptsize \textcolor{gray}{0.11}
 & 0.027 \scriptsize \textcolor{gray}{0.19}
 & 0.72 \scriptsize \textcolor{gray}{0.18} \\
\hline

\end{tabular}%
}
\end{table}

\noindent\textbf{Ablation studies:} To assess the design choices of our framework, we conducted extensive ablation experiments \(\mathcal{A}-\mathcal{E}\) on BrainMT for cognitive intelligence prediction, summarized in Table \ref{tab:ablation}. Exp \(\mathcal{A}\) shows that our choice of \(T=200\) is optimal, likely because fewer frames yield an insufficient signal-to-noise ratio, whereas more frames increase the risk of overfitting. Exp \(\mathcal{B}\) validates our hybrid design, where convolution extracts local spatial features, Mamba captures sequential dynamics, and transformers encode global context. Exp \(\mathcal{C}\) finds 12 Mamba layers and 8 transformers to be the best trade-off between complexity and capacity. Exp \(\mathcal{D}\) shows that replacing the bi-directional vision Mamba block with alternatives (e.g., VMamba \cite{liu2024vmamba}, MambaVision \cite{hatamizadeh2024mambavision}) degrades performance, emphasizing the importance of causal convolution for preserving fMRI’s temporal order. Finally, Exp \(\mathcal{E}\) demonstrates that our model surpasses SwiFT in subject-level Pearson correlations by processing more frames at once, capturing sequential brain dynamics more effectively. \\

\begin{figure}
    \centering
    \includegraphics[width=\linewidth]{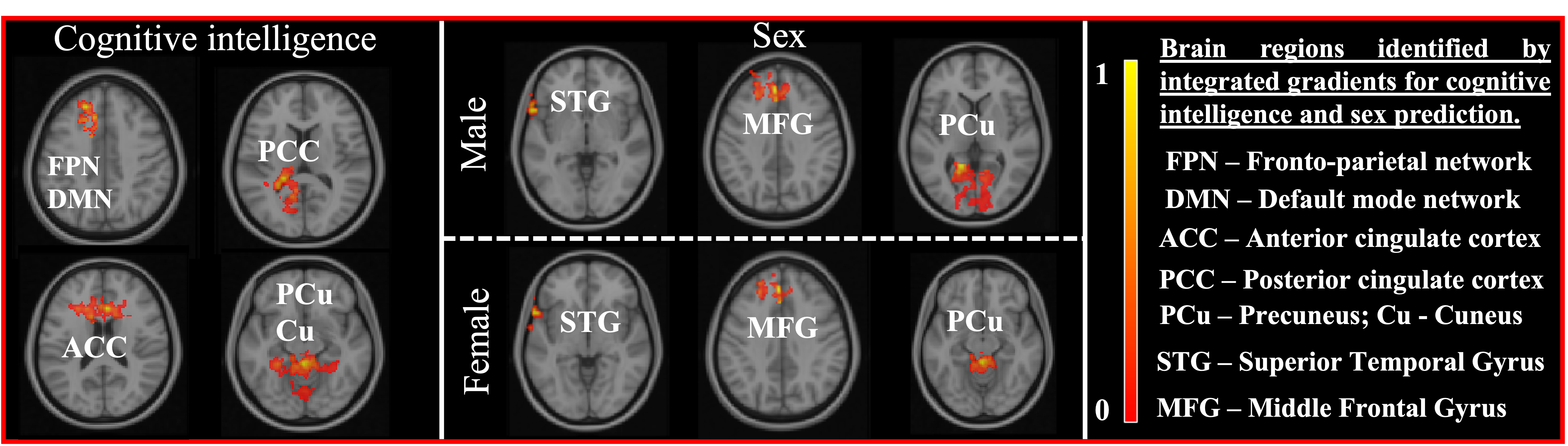}
    \caption{Consistent interpretability maps from BrainMT model, highlighting regions identified by Integrated Gradients for predicting cognitive intelligence and sex in HCP.}
    \label{fig:interpretation}
\end{figure}

\noindent\textbf{Clinical biomarkers:} To identify the brain biomarkers underlying BrainMT's phenotype predictions, we applied the Integrated Gradients (IG) algorithm \cite{sundararajan2017axiomatic} to assign importance scores to input features. As shown in Figure \ref{fig:interpretation}, the IG maps for cognitive intelligence reveal key contributions from regions within default mode network (DMN) and frontoparietal network (FPN), including posterior cingulate cortex (PCC), anterior cingulate cortex (ACC), precuneus (PCu), and cuneus (Cu). These areas are well-established in the literature for their roles in working memory, attention, decision-making, and visuospatial processing \cite{sestieri2011episodic,menon2011large}. For sex prediction, the IG maps consistently highlight superior temporal gyrus (STG), middle frontal gyrus (MFG), and PCu - findings that align with previous neuroimaging studies on sex differences \cite{ryali2024deep,weis2020sex}.

\vspace{-3mm}
\vspace{-5mm}

\section{Conclusion}

In this work, we introduced BrainMT, a novel hybrid deep network that captures the complex long-range temporal and spatial dependencies in volumetric fMRI data. By integrating a bi-directional Mamba block and a lightweight transformer block, BrainMT efficiently handles high-dimensional time series data and facilitates subject-level phenotypic characterizations. To the best of our knowledge, this is the first hybrid approach combining Mamba and transformer modules to manage extensive temporal coverage in volumetric rs-fMRI. This framework could serve as a general approach for prediction based on fMRI. In future work, we plan to incorporate self-supervised training on large-scale neuroimaging datasets to learn robust representations and further evaluate the model’s generalizability to downstream tasks with limited data.

\begin{credits}
\subsubsection{Acknowledgments.} This work was supported by the National Institutes of Health grants R01 EB029977 (PI Caffo) and R01 EB026549 (PI Lindquist) from the National Institute of Biomedical Imaging and Bioengineering. This work used data from the Human Connectome Project, WU-Minn Consortium (PIs David Van Essen and Kamil Ugurbil; NIH grant 1U54 MH091657, funded by the 16 NIH Institutes and Centers supporting the NIH Blueprint for Neuroscience Research and the McDonnell Center for Systems Neuroscience at Washington University), as well as from UK Biobank (Project ID 33278), a major biomedical database. This version of the contribution has been accepted for publication after peer review, but is not the Version of Record and does not reflect post-acceptance improvements or any corrections.

\subsubsection{Disclosure of Interests.}
The authors have no competing interests to declare that are relevant to the content of this article. 
\end{credits}
%
%
%
\bibliographystyle{splncs04}
\bibliography{mybibliography}
\end{document}